\documentclass[sigconf]{acmart}
\usepackage[normalem]{ulem}
\usepackage{multirow} 
\usepackage[section]{placeins}
\useunder{\uline}{\ul}{}
\AtBeginDocument{%
  }

\setcopyright{acmlicensed}
\copyrightyear{2018}
\acmYear{2018}
\acmDOI{XXXXXXX.XXXXXXX}

\acmConference[Conference acronym 'XX]{Make sure to enter the correct
  conference title from your rights confirmation emai}{June 03--05,
  2018}{Woodstock, NY}
\acmISBN{978-1-4503-XXXX-X/18/06}

\begin{document}

\title{Multi-Treatment Multi-Task Uplift Modeling for Enhancing User Growth}

\author{Yuxiang Wei}
\authornote{Work done as an intern in IEG, Tencent}
\email{weiyuxiang@gatech.edu}
\orcid{0000-0001-6552-8912}
\affiliation{%
  \institution{Georgia Institute of Technology}
  \city{Atlanta}
  \state{Georgia}
  \country{USA}
}

\author{Zhaoxin Qiu}
\affiliation{%
  \institution{ITE, Tencent}
  \city{Shenzhen}
  \country{China}}
\email{zhaoxinqiu@tencent.com}

\author{Yingjie Li}
\affiliation{%
 \institution{ITE, Tencent}
  \city{Shenzhen}
  \country{China}}
\email{wallaceyjli@tencent.com}

\author{Yuke Sun}
\affiliation{%
 \institution{ITE, Tencent}
  \city{Shenzhen}
  \country{China}}
\email{yukesun@tencent.com}

\author{Xiaoling Li}
\affiliation{%
 \institution{ITE, Tencent}
  \city{Shenzhen}
  \country{China}}
\email{serlinli@tencent.com}

\renewcommand{\shortauthors}{Trovato et al.}

\begin{abstract}
  As a key component in boosting online user growth, uplift modeling aims to measure individual user responses (e.g., whether to play the game) to various treatments, such as gaming bonuses, thereby enhancing business outcomes. However, previous research typically considers a single-task, single-treatment setting, where only one treatment exists and the overall treatment effect is measured by a single type of user response. In this paper, we propose a Multi-Treatment Multi-Task (MTMT) uplift network to estimate treatment effects in a multi-task scenario. We identify the multi-treatment problem as a causal inference problem with a tiered response, comprising a base effect (from offering a treatment) and an incremental effect (from offering a specific type of treatment), where the base effect can be numerically much larger than the incremental effect. Specifically, MTMT separately encodes user features and treatments. The user feature encoder uses a multi-gate mixture of experts (MMOE) network to encode relevant user features, explicitly learning inter-task relations. The resultant embeddings are used to measure natural responses per task. Furthermore, we introduce a treatment-user feature interaction module to model correlations between each treatment and user feature. Consequently, we separately measure the base and incremental treatment effect for each task based on the produced treatment-aware representations. Experimental results based on an offline public dataset and an online proprietary dataset demonstrate the effectiveness of MTMT in single/multi-treatment and single/multi-task settings. Additionally, MTMT has been deployed in our gaming platform to improve user experience.
\end{abstract}



\maketitle

\section{Introduction}

To offer a better personalized experience and increase user engagement, online marketing platforms usually provide incentives such as advertisements \cite{lo2002true}, discounts \cite{gubela2017revenue}, and bonuses \cite{ai2022lbcf}. Although these incentives are crucial for generating additional revenue and activity, they are often costly, and individual users can have varied responses to different incentives. For example, some users will not play the next game without a bonus, while others will continue to play regardless. Consequently, accurately modeling individual users' responses and identifying the target user groups that are likely to be positively affected by incentives is essential for enhancing marketing benefits \cite{xu2022learning}.

One of the fundamental challenges to measuring the response is the existence of the counterfactual problem, where an individual is either treated (treatment group) or not treated (control group). Therefore, in the same context, we can not simultaneously observe a user's response to a certain incentive or no incentive. Such a problem can be referred to as causal inference \cite{yao2021survey}. To resolve this, uplift modeling \cite{gutierrez2017causal} has been proposed to estimate the individual treatment effect (ITE) (a.k.a. uplift) that describes how individual user responds to an incentive \cite{zhang2021unified}. 

The current uplift modeling frameworks predominantly concentrate on directly modeling the response functions of both treatment and control groups to infer counterfactual predictions. Among them, meta-learner-based methods leverage existing models to estimate the Individual Treatment Effect (ITE) of personalized treatments. For example, S-learner \cite{kunzel2019metalearners} estimates the conditional average outcome of treatment and control group, then calculates ITE through subtraction. Building on this, other two-step meta-learners were proposed with other additional operations, including X-Learner \cite{kunzel2019metalearners}, DR-Learner \cite{kennedy2023towards}, R-Learner \cite{nie2021quasi}, etc. Nevertheless, these methods are prone to be influenced by the sample imbalance between the treatment and control groups. Another line of work involves tree-based models, which divide the user population into sub-groups according to specific splitting criteria (e.g., sensitivity to the treatment) and predict the uplift on each leaf node. A notable example is the causal forest \cite{athey2016recursive}, which integrates multiple trees to estimate heterogeneous treatment effects. With advances in deep learning, numerous neural network-based models have been developed that learn embeddings from related features, thereby predicting the uplift more flexibly. Based on representation learning, such models either predict the treatment effects of treatment and control groups separately  \cite{shi2019adapting,shalit2017estimating,curth2021inductive,curth2021nonparametric,schwab2020learning,zhong2022descn}, or directly models uplifts from user, context, and treatment features \cite{ke2021addressing,liu2023explicit,huang2024entire}. In this work, we focus on neural network-based models and propose to directly model uplifts.

\begin{figure}[h]
\centerline{\includegraphics[width=0.95\columnwidth]{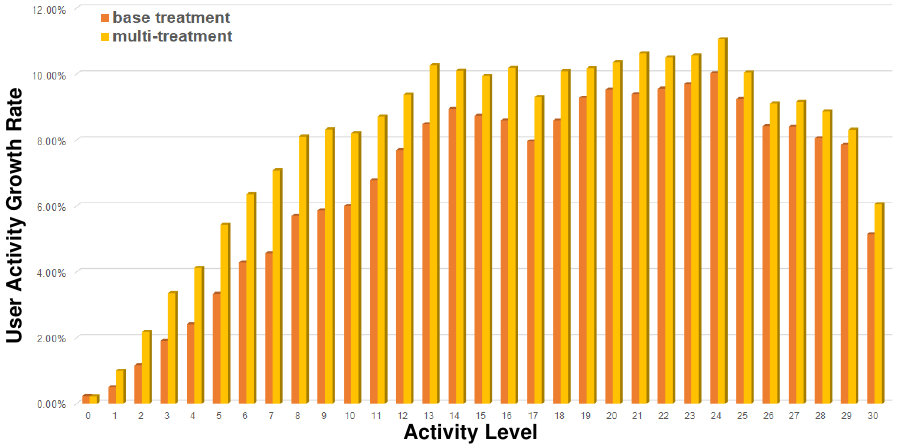}}
\caption{Activity growth rates by base treatment and multi-treatment deployment on our online gaming server. Growth rates are shown separately based on users' historical activity, as past activity significantly impacts future activity.}
\label{fig:baseincremental}
\end{figure}

While prevailing uplift models perform adequately on synthetic and product datasets, they exhibit two notable limitations. First, in real-world scenarios, multiple treatments often impact the target response, and multiple responses contribute to overall outcomes (multi-treatment multi-task). However, most models focus on a single treatment and a single target response (single-treatment single-task), overlooking the complex interactions between treatments and responses. This simplification can result in incomplete representations and biased ITE estimates. Several existing studies explore uplift modeling in multi-treatment or multi-task settings. For example, \cite{sun2024m,velasco2023hydranet} designed multi-head networks that predict the natural response (control group) and responses from multiple treatments, then calculate uplifts through subtraction. Addressing the multi-task problem, \cite{huang2024entire} studied two chained tasks (click-through rate and click-conversion rate) by designing a two-branch encoder with shared parameters between branches to encode features, subsequently outputting uplift scores for each task. Additionally, \cite{liu2023explicit} considered the single-task multi-treatment problem by designing separate encoders to encode treatment and user features, combining them to learn a unified representation, and then calculating the uplifts for each treatment. Nevertheless, in online applications, the response difference among treatments can be much less notable than the response difference between treatment and control groups. We illustrate this point in Fig. \ref{fig:baseincremental}. Therefore, simply combining the treatments and estimating the uplift can lead to suboptimal decisions. 

Secondly, existing models commonly focus on utilizing user and contextual features while neglecting the treatments. However, the correlation between treatments and user profiles is crucial for uplift modeling, especially in a multi-treatment setting. For instance, we observe that low-active users are more likely to play the next game when given a higher-valued bonus, whereas high-active users may continue to play the game regardless of receiving a bonus. While several works have incorporated treatment features as input to enhance the accuracy of estimation, they fall short in modeling the interaction among treatments and user features in a task-oriented way. For example, \cite{liu2023explicit,huang2024entire,ke2021addressing,xu2022learning} separately encoded the treatment features and non-treatment features, then generated a unified embedding either by concatenating or by weighted addition. This operation indirectly combines the treatments with other features but overlooks the implicit relationship between treatments and various tasks.

To address the aforementioned limitations, we propose a Multi-Treatment Multi-Task (MTMT) uplift modeling framework that directly models the user-treatment interactions. We identify multi-treatment as a tiered treatment problem, where the base treatment defines whether a user receives treatments, and the secondary treatments define the specific types and amounts. For instance, one must first decide whether to give a user a game bonus (which we refer to as \textit{Treatment Decision}) and then decide which bonus to give (which we refer to as \textit{Treatment Selection}). Following this, we separately estimate the base uplift for the treatment group and the incremental uplifts on top of the base uplift for the specific treatments. MTMT encodes the user features through a representation network based on a multi-gate mixture-of-experts. The generated embeddings are projected to compute the natural response for each task. Meanwhile, treatment encoders are employed to independently encode the base treatment and its subsequent secondary treatment information. Next, a treatment-user interaction module explicitly models how treatments attend to each user feature. The combined information is further enhanced and projected to compute the response of different treatment groups. This approach ensures a more accurate and nuanced estimation of treatment effects, addressing the complexities of multi-treatment and multi-task scenarios. In summary, our contributions are as follows:
\begin{itemize}
\item {\texttt{Background}}: We aim to accurately model the uplifts with multiple potential treatments and multiple target responses. To the best of our knowledge, this is the first effort in multi-treatment multi-task uplift modeling without underlying assumptions about the treatments or tasks.
\item {\texttt{Method}}: We introduce a novel uplift model that explicitly captures user-treatment feature interaction through the self-attention mechanism. By separately estimating the base uplifts and the uplift differences between treatments, the model accounts for the minimal uplift variance among treatments, thereby accurately estimating uplifts of different treatments and tasks in an end-to-end manner.
\item {\texttt{Evaluation}}: We demonstrate the effectiveness of the proposed model using a public dataset and a large-scale product dataset with multiple tasks and treatments. Our results indicate that MTMT significantly outperforms its competitors. The MTMT model has been deployed on our online gaming platform, serving millions of users.
\end{itemize}

\section{Related Works}

\subsection{Uplift Modeling}
Uplift modeling aims to establish a difference in the users' behaviors when applying or not applying certain treatments by measuring the corresponding ITE. Existing uplift research mainly focuses on three settings: 1) single-treatment single-task setting. For example, meta-learner-based methods \cite{kunzel2019metalearners,kennedy2023towards,nie2021quasi} integrate existing models to predict ITE. Besides this, tree-based methods \cite{wager2018estimation,athey2016recursive} gradually divide subpopulations by different metrics and estimate the ITE at the leaf node. Due to their superior feature extraction abilities, deep neural networks have gained much popularity in uplift modeling. They either directly model the uplift from learned representations \cite{ke2021addressing}, or separately estimate the natural response and treated response, then calculate the uplift by subtracting \cite{shalit2017estimating,shi2019adapting,curth2021inductive,gutierrez2017causal,schwab2020learning,zhong2022descn}. 2) single-treatment multi-task setting. Huang et al. \cite{huang2024entire} interpret the user behaviors as a sequential chain, and include two chained tasks to estimate their uplifts. They first extract user and treatment features with an encoder, then design a network with two branches that have shared parameters to predict the uplifts of the two tasks. 3) multi-treatment single-tasks setting. Zhao et al. \cite{zhao2019uplift} extend several meta-learners to the multi-treatment setting. Other works adapt tree-based methods to estimate multiple treatments' uplifts \cite{zhao2017uplift,zhao2017practically}. For the neural network-based approach, \cite{sun2024m,velasco2023hydranet,mondal2022memento} design independent heads to estimate each treatment group's response and the control group's response. Liu et al. \cite{liu2023explicit} employ the treatments as the input and explicitly extract its features, then predict the uplifts of each treatment through a single head. This paper considers the multi-treatment multi-task uplift modeling and explores its application in our online gaming server.

\subsection{Multi-task Learning}
Multi-task learning has been widely applied in computer vision, recommendation systems, and other fields. To boost the online business, multiple targets often influence marketing strategies. While focusing on a single task, existing models tend to ignore useful information from the training signals of related tasks \cite{ruder2017overview}. The introduction of multiple tasks mitigates the sample bias, where tasks with more training samples and be informative to tasks with fewer samples \cite{zhang2023advances}. In addition, training a multi-task model can reduce the cost of maintaining several models for each task. In the era of deep learning, a popular research line of multi-task learning is parameter sharing, where parameters are shared between different tasks. This includes hard parameter sharing \cite{guo2020learning}, which encodes representations of tasks into a shared embedding, and then applies task-specific heads to predict the outcome of each task. Extended from this, soft parameter sharing \cite{duong2015low} applies separate branches to model tasks and share information between branches by weighted addition or attention. Additionally, there is expert sharing \cite{ma2018modeling}, which utilizes several expert models to embed features and weigh the influence of each sub-task, and subsequently apply weights to calibrate the output of each expert. 

While proven effective in other domains, research on multi-task uplift modeling is still limited. We adopt the multi-gate mixture-of-expert in our framework to handle multiple tasks and individually estimate uplifts on each task.

\section{Problem Definition}
Assume the observed data to be $\mathcal{D} = \{[x_i,(\hat{t_i},t_i),y_i]\}_{i=1}^n$, where $x_i \in \mathbb{R}^{d}$ is the $d$-dimensional user features, $\hat{t_i} \in \{0, 1\}$ is the base treatment that denotes whether offering incentives to users, $t_i \in \mathbb{R}^{m}$ is the secondary treatment features with $m$ treatments, $y_i \in \mathbb{R}^{k}$ is the $k$ tasks, and $n$ is the number of samples.

We follow the Neyman-Rubin causal inference framework \cite{rubin2005causal} to define the estimation of ITE. Let $y_i^k(0)$ denotes the potential outcome of the $i$-th user in the control group on task $k$, and $y_i^k(m)$ denotes the outcome of receiving treatment $m$. Due to the counterfactual problem, we can only observe the outcome of a user in one treatment group, including the control group. As a result, there is no ground truth for supervised training. Instead, we estimate the expected response difference between the corresponding treatment and control. Let $\hat{\tau}^k(x_i)$ be $i$-th user's uplift when receiving a treatment, $\tau_m^k(x_i)$ be the incremental uplift under the $m$-th treatment and $k$-th task. Then the overall treatment effect can be computed as:
\begin{equation}
\begin{aligned}
    &\Gamma_m^k(x_i) =  \hat{\tau}^k(x_i) + \tau_m^k(x_i) \times \{\hat{t} = 1\} \\
    &=\mathbb{E}(y_i^k(1) - y_i^k(0) | x_i) + \mathbb{E}(y_i^k(m) - y_i^k(1) | x_i, \hat{t} = 1)
\end{aligned}
\label{eqUplift}\end{equation}
where $\Gamma_m^k(x_i)$ is the $i$-th user's overall uplift score on the $k$-th task, $\hat{t}$ signifies if an individual is in the treatment group.

\section{Methodology}

\begin{figure*}[h]
\centering
\includegraphics[width=0.95\textwidth]{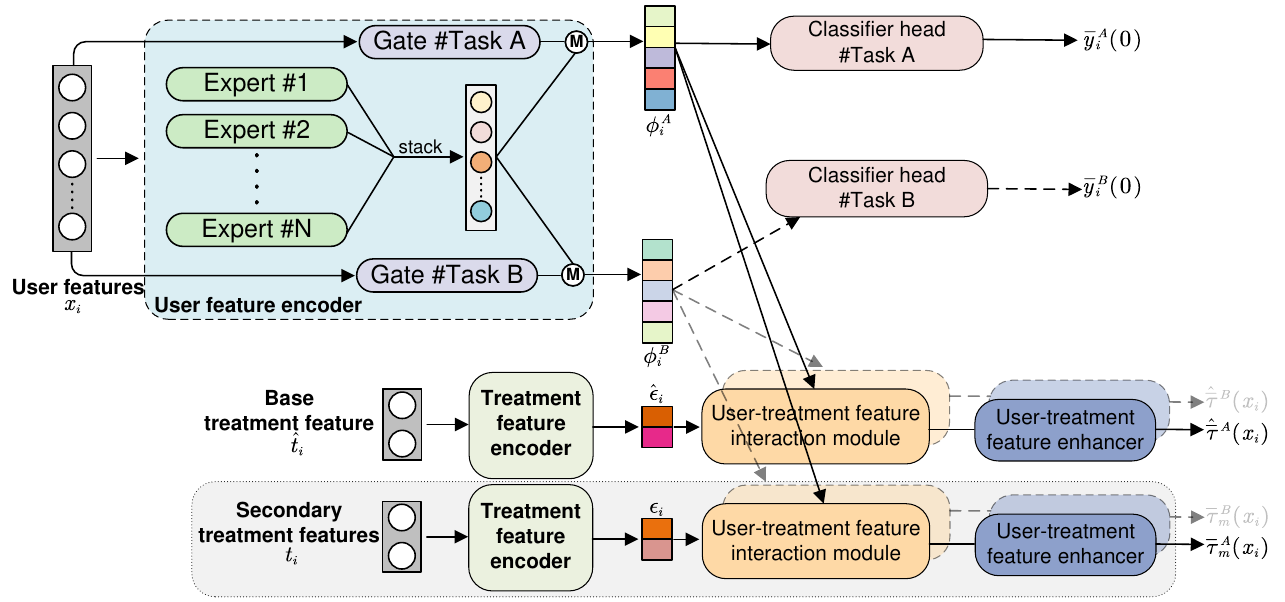}
\caption{Illustration of the proposed \textbf{M}ulti-\textbf{t}reatment \textbf{M}ulti-\textbf{t}ask (MTMT) framework. Note that for clarity we only show the network structure of two tasks.}
\label{fig:framework}
\end{figure*}

\subsection{Architecture}
The illustration of the proposed multi-treatment multi-task (MTMT) uplift modeling network is shown in Fig. \ref{fig:framework}. Given a sample $\{x_i,(\hat{t_i},t_i),y_i\}$, the user features $x_i$ are first encoded by the user feature encoder to generate the representations for each task $\{\phi^0, \phi^1, \cdots, \phi^k \}$. Meanwhile, the base treatment feature $\hat{t_i}$ (indicates control or treatment) and the secondary treatment features $t_i$ (indicates specific types of treatments) are encoded separately to generate the corresponding representations $\hat{\epsilon}$ and $\epsilon$. The feature representations of each task are projected by the corresponding classifier head to compute its natural response $\bar{y}_i^k(0)$ of task $k$ when not treated. Additionally, the embeddings of base and secondary treatments are fed separately into the corresponding user-treatment interaction module, in which the cross-correlations between treatment and user features are computed. The resultant treatment-aware features are further enhanced and used to estimate the response difference for receiving treatment and the incremental uplift for receiving treatment $m$.

\subsection{Task-Oriented Feature Encoder}
To explicitly model inter-task relationships and learn task-specific representations, we adopt the multi-gate mixture-of-experts (MMOE) \cite{ma2018modeling} as the user feature encoder. Mixture-of-experts is a form of ensemble learning that integrates numerous expert models (e.g., vanilla CNN) to learn a shared representation and use the combined predictions to improve accuracy \cite{eigen2013learning}. Extended from this, MMOE introduces an additional gating network to filter useful information from the shared representation of each task. To obtain the encoded representation $\phi_i^k$ for task $k$:
\begin{equation}
\begin{aligned}
    & \phi_i^k = \mathcal{G}^k(x_i) \cdot \{\mathcal{E}^1(x_i), \mathcal{E}^2(x_i), \cdots, \mathcal{E}^n(x_i) \}
\end{aligned}
\label{eqMMOE}\end{equation}
where $\mathcal{E}^j$ is the $j$-th expert network and there are $n$ experts and $\mathcal{G}^k$ is the gating network for $k$-th task. We choose ResNet18 \cite{he2016deep} as the backbone for each expert. Note that the embeddings from the experts are stacked and then filtered by the corresponding gate to produce a task's representation. The gating network can be a simple linear projection from the input with additional activation:

\begin{equation}
    \mathcal{G}^k = \mathrm{softmax}(W_g^kx_i)
\label{eqGate}\end{equation}
where $W_g^k \in \mathbb{R}^{n \times d}$ is the trainable weights, $n$ is the number of experts, and $d$ is the feature dimension.

The representations $\phi^k_i$ for each task only contain non-treatment information and are therefore used to estimate the natural response of the control groups by a linear projection head:
\begin{equation}
    \bar{y}_i^k(0) = W^{\mathrm{proj}}_i\phi_i^k
\end{equation}

Since most treatments are binary or discrete, we first one-hot encode the base and secondary treatment seprately and multiply the resultant sparse vector with a corresponding learnable dense matrix $A_i \in \mathbb{R}^{v \times m}$ to produce embeddings:
\begin{equation}
    \hat{\epsilon}_i = \hat{A}_i \hat{t}_i,\quad \epsilon_i = A_i t_i
\end{equation}
where $v$ is the embedding dimension and $m$ is the number of possible treatments.

It should be noted that both user and treatment feature encoders can be substituted with other feature representation learning networks, provided they have the appropriate dimensions. For instance, a single ResNet18 can be used for user feature extraction, adapting the Multi-Task Multi-Treatment (MTMT) model to a single-task setting. The flexibility of the proposed design allows for seamless online implementation across various problem settings.

\subsection{User-Treatment Feature Interaction}
\begin{figure}[h]
\centerline{\includegraphics[width=0.95\columnwidth]{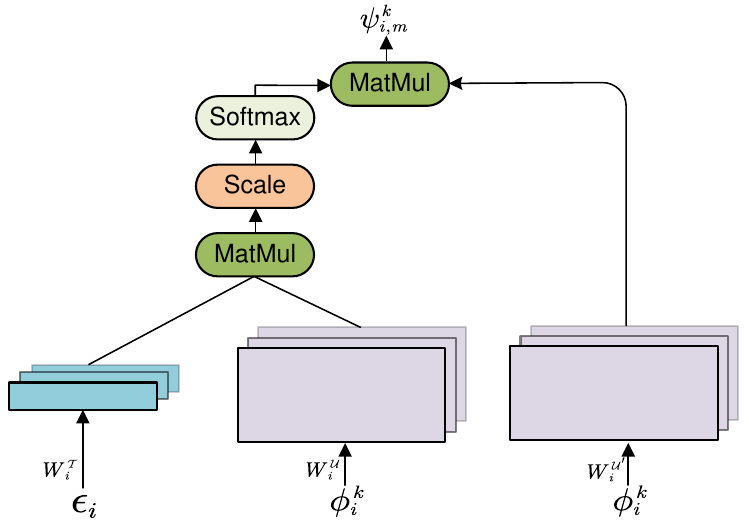}}
\caption{The proposed user-treatment feature interaction module}
\label{fig:interact}
\end{figure}

To explicitly utilize treatment features and model their relationships with user features, we propose the user-treatment feature interaction module based on self-attention \cite{vaswani2017attention}. The structure is shown in Fig. \ref{fig:interact}. We treat the treatment embeddings $\epsilon_i$ as the query and the user feature embeddings $\phi_i^k$ as the key and value. Subsequently, we compute how each treatment attends to each user feature, then use the resultant attention scores to generate treatment-aware embeddings:

\begin{equation}
    \psi^k_{i,m} = \mathrm{softmax}(\frac{W^\mathcal{T}_i \epsilon_i \times (W^\mathcal{U}_i \phi^k_i)^T}{\sqrt{d_\mathcal{U}}}) W_i^{\mathcal{U}^\prime} \phi^k_i
\end{equation}
where $W^\mathcal{T}_i$ linearly projects the treatment embedding and $W^\mathcal{U}_i$, $W_i^{\mathcal{U}^\prime}$ linearly projects the user feature embeddings. $\sqrt{d_\mathcal{U}}$ is a scaling factor. We then process $\psi^k_i$ by a user-treatment feature enhancer to further refine the useful information and the resultant embeddings are projected to estimate the base uplift score $\hat{\bar{\tau}}^k(x_i)$ and the incremental uplift score $\bar{\tau}^k_m(x_i)$ for treatment $m$:

\begin{equation}
    \hat{\bar{\tau}}^k(x_i) = \hat{W}^{\tau}_i \mathrm{MLP(}\bar{\psi}^k_{i,m}), 
    \quad\bar{\tau}^k_m(x_i) = W^{\tau}_i \mathrm{MLP}(\psi^k_{i, m})
\label{eqTau}\end{equation}

Here we use a multi-layer perception (MLP) as the feature enhancer and $W^{\tau}_i$ is the projection matrix.

\subsection{Multi-Treatmet Multi-Task Uplift Estimation}
Given the natural response $\bar{y}_i^k(0)$, the base uplift score $\hat{\bar{\tau}}^k(x_i)$, and the incremental uplift score $\bar{\tau}^k_m(x_i)$, we can estimate the user's response by:
\begin{equation}
    \bar{y}_i^k(m) = \bar{y}_i^k(0) + \hat{\bar{\tau}}^k(x_i) + \bar{\tau}^k_m(x_i)
\label{eqYk}\end{equation}

Here $\hat{\bar{\tau}}^k(x_i)$ indicates how the user acts to a treatment, and $\bar{\tau}^k_m(x_i)$ indicates the incremental effect of the specific treatment $m$, on top of the base uplift. Intuitively, $\bar{\tau}^k_m(x_i)$ is only effective when the user is in the treatment group. For online deployment when the treatment information is unknown, we permute all possible combinations of treatments and rank the resultant base and incremental uplifts.

Compared to previous methods for multi-treatment uplift modeling, the proposed approach separately estimates the effects of "whether to treat" and “what to treat", thereby bridging the numerical gap between the base uplift and the uplift variations among treatments.

\subsection{Training and Inference}

We use the natural response and treated response for each task to compute the overall loss across the entire sample space:
\begin{equation}
\begin{aligned}
    & \sum_{k=1}^N [\sum_{i \in \mathrm{C}} \mathcal{L}(y_i, \bar{y}_i^k(0)) + \sum_{i \in \mathrm{T}} \mathcal{L}(y_i, \bar{y}_i^k(m))] \\
    &=\sum_{k=1}^N [\sum_{i \in \mathrm{C}} \mathcal{L}(y_i, \bar{y}_i^k(0)) + \sum_{i \in \mathrm{T}} \mathcal{L}(y_i, \bar{y}_i^k(0) + \hat{\bar{\tau}}^k(x_i) + \bar{\tau}^k_m(x_i))]\\
\end{aligned}
\label{eqLoss}\end{equation}
where $\mathcal{L}$ is the mean-squared error loss function, C denotes the control group, and T denotes the treatment group. At inference, we only use Eq.\ref{eqTau} to directly compute the base and incremental uplifts, which are subsequently ranked and used to determine whether to offer treatment and which treatment to offer.

\section{Experiments}

We conduct extensive experiments to verify the effectiveness of the proposed MTMT in single-treatment single-task, single-treatment multi-task, multi-treatment single-task, and multi-treatment multi-task settings. We mainly focus on the following questions:
    
\begin{itemize}
    \item \textbf{RQ1:} Can the proposed MTMT outperform other baseline methods on public and product datasets?
    \item \textbf{RQ2:} How does each design contribute to the overall performance of MTMT?
    \item \textbf{RQ3:} Whether the model produces interpretable results that are consistent with our online observations?
\end{itemize}

\subsection{Experimental Setup}
\subsubsection{Datasets}

\begin{itemize}
    \item \textbf{CRITEO} \cite{diemert2021large}: CRITEO is an open-sourced uplift modeling dataset for online advertising. The data is created by compiling data from various incremental tests, with a specific type of randomized trial in which a portion of the population is randomly excluded from advertising targeting. We include about 14 million samples, each has 12 continuous features, and use visit as the target.
    \item \textbf{Product}: We include a product dataset containing over 10 million samples collected from the our online gaming platform. There are about 700 discrete and continuous features that describe users' static profiles and their recent gaming histories. To minimize the influence of confounding factors in uplift modeling, we gather data from randomized controlled trials. In these trials, treatments are assigned randomly to ensure an even distribution of potential outcomes between the treatment and control groups. For the multi-task scenario, we employ two binary labels: whether the user logs in the next day (short-term activity) after receiving or not receiving the treatment and whether the user plays more games in the next 7 days (long-term activity). For the multi-treatment scenario, we identify the base treatment as "whether to give a bonus" and the secondary treatment as "the type of bonus" (bonus type A and bonus type B), both of which are binary-valued. Note that in the practical application, certain bonus types are only available for a subset of users. Therefore, we collect an additional multi-treatment dataset by selecting users who are accessible by all types of bonuses.
\end{itemize}

\subsubsection{Baselines}
To demonstrate the performance of MTMT, we include a set of popular methods proposed for uplift modeling, including: S-Learner \cite{kunzel2019metalearners}, T-Learner \cite{kunzel2019metalearners}, CFR \cite{shalit2017estimating}, DragonNet \cite{shi2019adapting}, EUEN \cite{ke2021addressing}, DESCN \cite{zhong2022descn}, FlexTENet \cite{curth2021inductive}, EFIN \cite{liu2023explicit}, and M3TN \cite{sun2024m}. 

For the multi-treatment problem, we employ EFIN, M3TN, and HydraNet \cite{velasco2023hydranet}. Furthermore, we extend S-Learner and T-Learner to handle multiple treatments. For S-Learner, we directly apply the multi-treatment features as the input and iterate all possible treatment assignments to compute the corresponding uplift score. For T-Learner, we use multiple branches to process each treatment individually.

\subsubsection{Evaluation Metrics}
Following the previous works, we adopt Area Under the QINI Curve (QINI), Area Under the Uplift Curve (AUUC), and the uplift score at first 30\% (LIFT@30) to evaluate the uplift ranking capability of different models. Note that for easier and fairer comparison, we employ the normalized QINI and AUUC.

\subsubsection{Implementation Details}
We train all models on NVIDIA A100, with Pytorch 2.1.2 and Python 3.11. We use the AdamW optimizer \cite{kingma2014adam} with a learning rate of 0.001 and cosine annealing learning rate scheduler \cite{loshchilov2016sgdr}. Additionally, we use a batch size of 15360 and set the maximum epochs as 50. For the detailed parameter setting of MTMT, we employ the standard ResNet without the classifier head as the expert in the user feature encoder and set the number of experts to 4. 

\subsection{RQ1: Performance Comparison}
We evaluate the single-treatment single-task variation of the proposed methods and the baselines for single-treatment and multi-treatment uplift estimation. The results are presented in Table \ref{tab:perfom} and Table \ref{tab:perfomMtreat}. Note that for MTMT in the single-treatment case, we only use base uplift $\hat{\bar{\tau}}^k(x_i)$ from Eq.\ref{eqTau} to decide whether a treatment should be offered.

In Table \ref{tab:perfom}, we test all models based on two tasks on the product dataset: short-term activity and long-term activity. We further show how the multi-task version of MTMT performs on the two tasks. From the table, MTMT has 0.164 QINI on the CRITEO dataset. Meanwhile, the best-performing baseline on the CRITEO dataset is FlexTENet, which reaches 0.0779 QINI. On the product dataset, MTMT still maintains its advantage over other baselines on both tasks. For the multi-task setting, MTMT's performance slightly degrades, while its overall performance is still much better than the baselines.

To further validate MTMT's performance in the multi-treatment setting, we present the results in Table \ref{tab:perfomMtreat}. We separately calculate the metrics on the two types of treatments, bonus A and bonus B. From the table, among the baselines, M3TN performs the best on bonus A and EFIN performs the best on bonus B, while MTMT outperforms all its comparatives on the two types of treatments.

\begin{table*}[t]
\centering
\begin{tabular}{c|ccc|ccc|ccc}
\hline
Dataset                    & \multicolumn{3}{c|}{CRITEO}                        & \multicolumn{3}{c|}{Product - short-term activity} & \multicolumn{3}{c}{Product - long-term activity}  \\ \hline
Metrics                    & QINI           & AUUC            & LIFT@30         & QINI            & AUUC           & LIFT@30         & QINI           & AUUC            & LIFT@30        \\ \hline
S-Learner                  & 0.0703         & 0.0283          & 0.0258          & 0.0420          & 0.0880         & 0.00575         & \textit{0.131}    & 0.0101          & 0.0873         \\
T-Learner                  & 0.0706         & 0.0286          & 0.0271          & 0.0421          & \textit{0.0890}   & 0.00763         & 0.111          & 0.00837         & 0.0831         \\
CFR                        & 0.0715         & 0.0295          & 0.0278          & 0.0100          & 0.0182         & 0.0037& 0.108          & \textit{0.0211}    & \textit{0.0915}   \\
DragonNet                  & 0.0183         & 0.00735         & 0.0121          & 0.00743         & 0.0156         & 0.00393& 0.0812         & 0.0164          & 0.0792         \\
EUEN                       & 0.0730         & 0.0297          & 0.0279          & 0.00827         & 0.0262         & 0.00313& 0.106          & 0.0113          & 0.0778         \\
DESCN                      & 0.0718         & 0.0289          & 0.0264          & 0.0351          & 0.0771         & 0.00438         & 0.0073& 0.0110& 0.0177\\
FlexTENet                  & \textit{0.0779}   & \textit{0.0322}    & \textit{0.0290}    & 0.0321          & 0.0684         & 0.00268         & 0.111          & 0.0106          & 0.0796         \\
EFIN                       & 0.0725         & 0.0293          & 0.0215          & \textit{0.0725}    & 0.0293         & \textit{0.0415}    & 0.0340& 0.0281& 0.0554\\
M3TN                       & 0.0395         & 0.0176          & 0.0205          & 0.0295          & 0.0416         & 0.00260& 0.108          & 0.0204          & 0.0908         \\ \hline
\textbf{MTMT}              & \textbf{0.164} & \textbf{0.0593} & \textbf{0.0338} & \textbf{0.0886} & \textbf{0.155} & \textbf{0.0638} & \textbf{0.360} & \textbf{0.0579} & \textbf{0.110} \\ \hline
\textbf{MTMT (multi-task)} & \textbf{--}    & \textbf{--}     & \textbf{--}     & \textbf{0.0586} & \textbf{0.118} & \textbf{0.0326} & \textbf{0.154} & \textbf{0.0218} & \textbf{0.111}
\end{tabular}
  \caption{Overall performances of the single-treatment single-task version of MTMT and its comparatives on the public and product datasets. We use two targets of the product dataset, with a single treatment (whether to offer a bonus). Note that the last row is the multi-task version of MTMT. The best baselines are tilted and the best methods are marked as bold for each metric.}  \label{tab:perfom}
\end{table*}

\begin{table*}[t]
\centering
\begin{tabular}{c|ccc|ccc}
Treatment                       & \multicolumn{3}{c|}{Bonus A}                        & \multicolumn{3}{c}{Bonus B}                         \\ \hline
Metrics                         & QINI            & AUUC            & UPLIFT@30       & QINI            & AUUC            & UPLIFT@30       \\ \hline
S-Learner                       & 0.0126& 0.0236& 0.00211& 0.00771         & 0.0152          & 0.00105         \\
T-Learner                       & 0.00505         & 0.00878         & 0.00716& 0.00374         & 0.00942         & 0.00141\\
HydraNet                        & 0.00456         & 0.00844         & 0.0182& 0.00616         & 0.012           & 0.00895\\
EFIN                            & 0.0142          & 0.0256          & \textit{0.00766}   & \textit{0.0294}    & \textit{0.0513}    & \textit{0.0334}    \\
M3TN                            & \textit{0.0162}    & \textit{0.0281}    & 0.00259         & 0.00107         & 0.00640& 0.0175\\ \hline
\textbf{MTMT (multi-treatment)} & \textbf{0.0324} & \textbf{0.0653} & \textbf{0.0344} & \textbf{0.0291} & \textbf{0.0782} & \textbf{0.0491}
\end{tabular}
\caption{Overall performances of MTMT and its comparatives on the multi-treatment product datasets. The best baselines are tilted and the best methods are marked as bold for each metric.}  \label{tab:perfomMtreat}
\end{table*}

\subsection{RQ2: Ablation Study}
We conduct ablation studies to validate the effectiveness of the key design of MTMT. Specifically, for the architecture, we remove the user-treatment feature enhancer and substitute the user-treatment feature interaction module with matrix multiplication. For the multi-treatment case, we remove the "secondary treatment features" branch as in Fig.\ref{fig:framework} and estimate uplifts of multiple treatments with a single output. For the multi-task case, instead of estimating ITE for each task, we jointly estimate the ITE for all tasks. The results are presented in Table\ref{tab:ablation}. From the table, changing the network architecture or modifying the ITE estimation process can result in performance degradation.

\begin{table*}[h!]
\centering
\begin{tabular}{c|c|ccc}
\hline
                              &                                           & \multicolumn{3}{c}{Product}                                              \\ \hline
                              &                                           & QINI                   & AUUC                   & LIFT@30                \\ \hline
\multirow{2}{*}{architecture} & w/o user-treatment feature interaction    & 0.0426                 & 0.0881                 & 0.00849                \\
                              & w/o user-treatment feature enhancer       & 0.0759                       & 0.145                       & 0.0474                       \\
\multicolumn{1}{l|}{}         & \textbf{MTMT}                             & \textbf{0.0886}        & \textbf{0.155}         & \textbf{0.0638}        \\ \hline
multi-treatment               & w/o tiered treatment effect estimation    & 0.0249/0.0233          & 0.0464/0.0514          & 0.0164/0.0186          \\
                              & \textbf{MTMT (multi-treatment)}           & \textbf{0.0324/0.0291} & \textbf{0.0653/0.0782} & \textbf{0.0344/0.0491} \\ \hline
multi-task                    & w/o task-wise treatment effect estimation & 0.0552/0.108           & 0.113/0.0118           & 0.035/0.0922           \\ 
                              & \textbf{MTMT (multi-task)}                & \textbf{0.0586/0.154}  & \textbf{0.118/0.0218}  & \textbf{0.0326/0.111} 
\end{tabular}
\caption{Ablation study of MTMT on the product dataset. For the multi-treatment setting, we show the model's performance on different treatments as "bonus A/bonus B". For the multi-task setting, we show the model's performance on different tasks as "short-term activity/long-term activity". The proposed designs are marked in bold.}  \label{tab:ablation}
\end{table*}

\subsection{RQ3: Interpretable Analysis}
\begin{figure}[h]
\centerline{\includegraphics[width=\columnwidth]{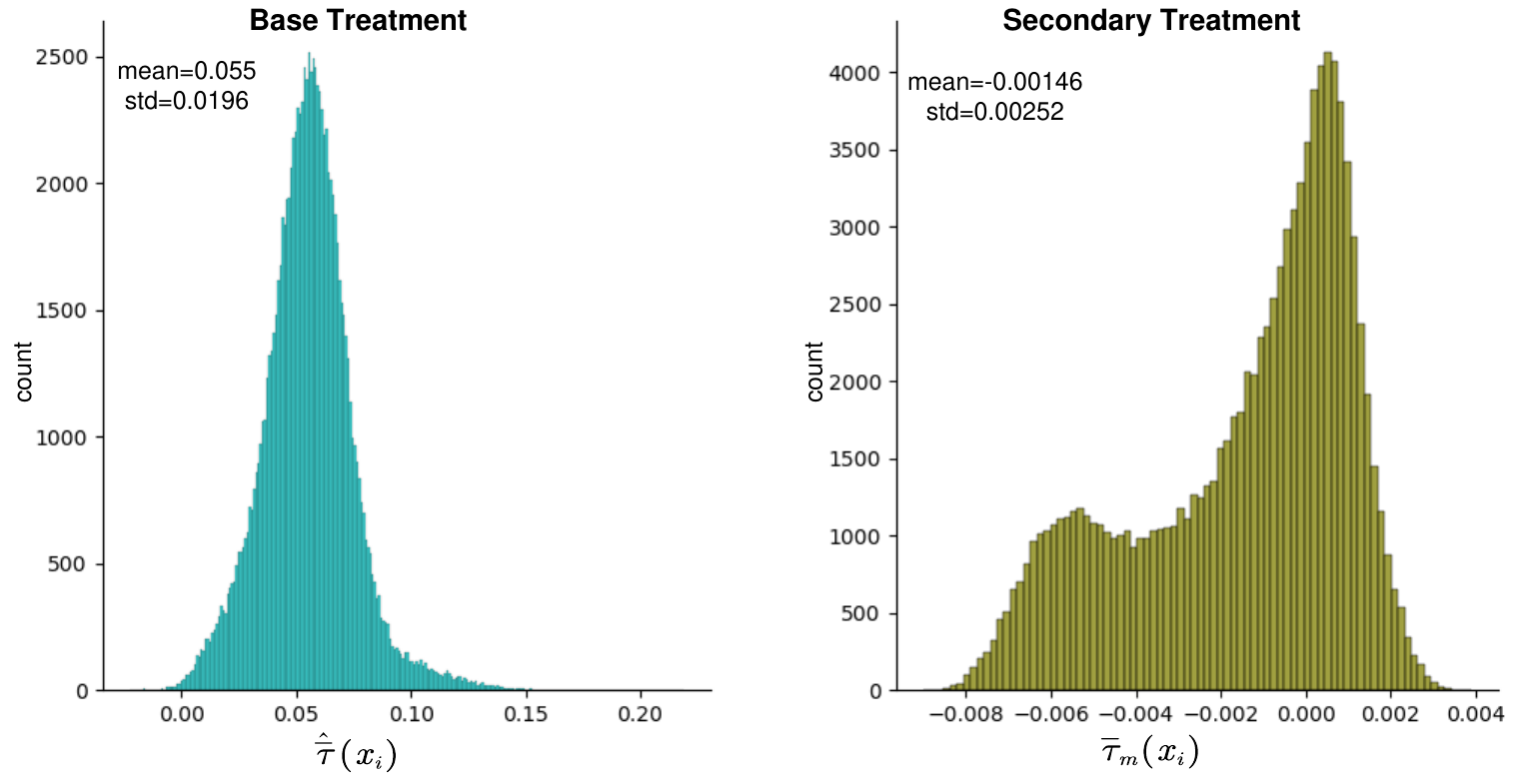}}
\caption{Distributions of base and incremental treatment effects. The base treatment effects are numerically much larger than the incremental treatment effects.}
\label{fig:dist}
\end{figure}

To demonstrate the model produces interpretable results that are consistent with our online observations, we visualize the density distributions of the base treatment effect ($\hat{\bar{\tau}}^k(x_i)$) and the incremental treatment effect ($\bar{\tau}^k_m(x_i)$) using our product dataset. As shown in Fig. \ref{fig:dist}, the average of $\hat{\bar{\tau}}^k(x_i)$ (0.055) is much larger than the average of $\bar{\tau}^k_m(x_i)$, which corresponds to our observations in the online gaming server (Fig. \ref{fig:baseincremental}).

To demonstrate the effectiveness of the proposed user-treatment interaction module, we visualize the generated correlations between user features and treatments. The illustration can be found in the first section of appendix.

\subsection{Online Deployment}
\begin{figure}[h!]
\centerline{\includegraphics[width=0.95\columnwidth]{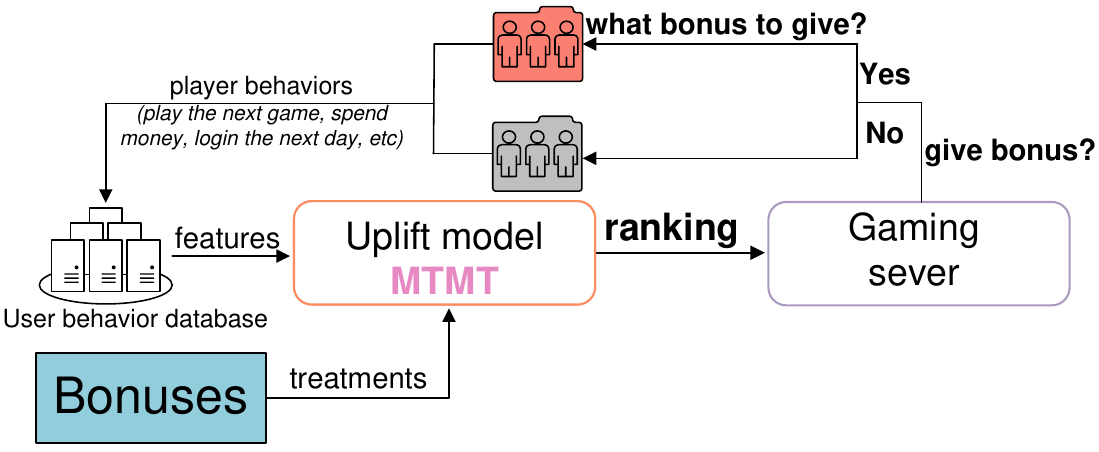}}
\caption{Overview of our online bonus deployment platform}
\label{fig:process}
\end{figure}

We aim to apply uplift modeling to personalize the distribution of bonuses to enhance the flow experience for users on our online gaming platform. Different bonuses are offered to users to ensure a better gaming experience, thereby motivating user activity and willingness to spend money. The online deployment process is illustrated in Fig. \ref{fig:process}. After training MTMT offline with treatment and user features, the model ranks users into buckets based on their estimated base and treatment effects. For each bucket, the gaming server determines whether to offer a user a bonus and, if so, which bonus to offer, based on certain constraints.

\section{Conclusion and Future Work}
In this paper, we explored uplift modeling within a multi-treatment, multi-task framework for an online gaming platform. To effectively extract features for different tasks and accurately estimate the effects of various treatments, we proposed the Multi-Treatment Multi-Task (MTMT) uplift modeling framework that explicitly utilizes user-treatment features to estimate the task-wise uplifts. Additionally, to precisely capture the uplift differences between various treatments, we proposed separately estimating the base treatment effect and the incremental treatment effect. The base treatment effect pertains to whether a treatment should be given, while the incremental treatment effect addresses the specific type of treatment. Extensive experiments and ablation studies validated the effectiveness of our methods. We also provided interpretable analyses to demonstrate how user features correlate with treatments. In future work, we plan to extend our models to accommodate non-binary treatments and non-binary tasks, thereby increasing their applicability to a broader range of scenarios.

\bibliographystyle{ACM-Reference-Format}
\bibliography{sample-base}

\end{document}